\documentclass{llncs}
\usepackage{graphicx, fixltx2e, setspace}
\usepackage{amsmath, amsfonts, amssymb}
\usepackage{graphicx, color, import}
\usepackage{latexsym}
\usepackage{caption}
\usepackage{braket}
\usepackage{tikz}
\usepackage{natbib}
\usetikzlibrary{positioning, arrows, shapes}

\begin{document}
\title{A Compositional Explanation of the `Pet Fish' Phenomenon}
\date{\today}
\author{Bob Coecke\inst{1} \and Martha Lewis\inst{1}}
\institute{Department of Computer Science, University of Oxford\\ Parks Rd, Oxford, OX1 3QD}
\maketitle

\begin{abstract}
The `pet fish' phenomenon is often cited as a paradigm example of the `non-compositionality' of human concept use. We show here how this phenomenon is naturally accommodated within a compositional distributional model of meaning. This model describes the meaning of a composite concept by accounting for interaction between its constituents via their grammatical roles. We give two illustrative examples to show how the qualitative phenomena are exhibited. We go on to apply the model to experimental data, and finally discuss extensions of the formalism.
\end{abstract}

\section{Introduction}
\label{sect:intro}
A question in cognitive science is to characterise how humans understand expressions involving the composition of many concepts. The principle of semantic compositionality is often attributed to Frege, and is summarized by \cite{kp} as follows:
\begin{quote}
The meaning of a complex expression is a function of the meaning of its parts and of their syntactic mode of combination.
\end{quote}
One approach to understanding the meaning of a complex expression is to model concepts as sets, and the meaning of a complex expression as determined by various operations on its parts. Some combinations of concepts such as `red car' can be understood in this way, by taking the intersection of things that are red and things that are cars, whereas others such as `red wine' cannot be characterised as a straightforward intersection, because red wine is not a standard red colour. This failure of concepts to behave conjunctively is showcased via the `pet fish' problem, given in \cite{osh1981} as a counterexample to a fuzzy set model of concepts. The problem is as follows. Concepts, as used by humans, are characterised as fuzzy sets, with the typicality of an item $x$ to the concept $L$ given by a membership function $m_L(x)$.  A goldfish is considered not to be a very typical pet, nor a very typical fish, say $m_{pet}(goldfish) = m_{fish}(goldfish) = 0.5$. However the goldfish is a typical pet fish, say $m_{pet~fish}(goldfish) = 0.9$. This contradicts rules of conjunction in fuzzy set theory, in which the membership of an item in the intersection of two fuzzy sets is given by the minimum of its membership in either one. The phenomenon 
\[
m_{A\land B}(x) > \min (m_{A}(x), m_{B}(x))
\] is called \emph{overextension} and 
\[
m_{A\lor B}(x) < \max (m_{A}(x), m_{B}(x))
\] \emph{underextension}.

This and similar phenomena have been experimentally verified in \cite{hamp1988a, hamp1988b}. The data collected is a series of typicality ratings for concepts and their combinations. Further evidence is reported in \cite{hamp1987}, in which the importance of attributes in conjunctions of concepts is dependent on the importance of the same attributes in the conjuncts. In addition to the phenomena of over- and underextension, non-commutativity of membership in conjunctions is exhibited, together with attributes becoming more or less important depending on the order of combination. These phenomena are very puzzling if concepts are conceived of as fuzzy sets. \cite{kp} themselves provide a thorough analysis of these phenomena, and propose a model which circumvents some of these problems. However, it is not able to account for the `pet fish' phenomenon.

In order to solve the problems encountered by the fuzzy set theoretic view of concepts, quantum theory inspired models of concepts are developed in, amongst others,  \cite{aerts, aerts2, aerts2013, pothos2012}. These models view concepts as subspaces of a vector space. Similarity of one concept to another is measured by the projection of one concept onto another. The combination of two concepts is modelled as a tensor product. The `pet fish' phenomenon then arises when the representation of the combined concept is entangled. The combination of more than two concepts may also be effected simply by taking the tensor product of the constituent concept spaces. These approaches, however,  require that the representational space for a concept grows in size as more elements are added to the compound. This means that compounds will become unwieldy, and more importantly that compound concepts cannot be directly compared with their constituents. 

It is then possible to investigate how well these models characterise human concept combination by examining the probability of agreeing with certain statements, or rating similarity between concepts. A review of such phenomena is given in \cite{pothos2012}. Data such as that collected by \cite{hamp1988a, hamp1988b} is argued to provide evidence for quantum structure in human thought \citep{aerts2009}, since the similarity ratings do not adhere to classical probability theory.  This failure to adhere to classical probability theory is then termed `non-compositionality'. However, whilst these approaches can provide a modelling of the data, they do not give an explanation of the fact that humans \emph{do} successfully combine concepts. Further, the role of syntax is ignored. We will argue that the compositional nature of human thought can be explained, in a way that retains the qualitative phenomena discussed, and in a way that maps meanings to one shared space.

\cite{clark2007} represent the meaning of words as vectors, and use the tensor product to combine words and their grammatical roles into sentences, using ideas developed in \cite{smolensky2006}. However, this again has the limitation, explicitly recognised, that sentences and words of different compositional types may not be compared. \cite{coecke2010} in part inspired by the high-level description of quantum information flow introduced in \cite{Abramsky2004}, utilise grammar in order to use composite spaces without increase in size of the resulting meaning space. This is the model that we will use to explain these compositional phenomena. The meaning of a composite concept, or a sentence, arises from the meaning of its constituents, mediated by grammar. This allows composite concepts to be directly compared with their constituents, and the meaning of sentences of varying length and type to be compared. In this paper we will apply this compositional distributional model of meaning to the `pet fish' problem, and show how the qualitative phenomena of overextension and non-commutativity naturally arise - i.e., we show how a goldfish can be a good `pet fish', even though it is neither a typical pet, nor a typical fish, and within this we also explain why a `pet fish' is different to a `fishy pet'. We provide a modelling of experimental data, and go on to discuss how a fuller account of human concept use will be developed. 

The remainder of the paper is organised as follows. In section \ref{sec:disco}, we describe the compositional distributional model of semantics. In section \ref{sec:petfish} we describe how the `pet fish' phenomenon may be modelled and give two illustrative examples. We go on to model experimental data in section \ref{sec:exp}, and discuss the results and further work in section \ref{sec:disc}.

\section{A Compositional Distributional Approach to Meaning}
\label{sec:disco}
The account of meaning of \cite{coecke2010}, unifies compositional accounts of meaning, where the meaning of a sentence is seen as a function of the meanings of the individual words in the sentence, and distributional accounts of meaning, where the meaning of individual words are characterised as vectors. The main source of word meanings in distributional semantics is to derive word meanings from text corpora via word co-occurrence statistics and a review of techniques. Other methods for deriving such meanings may be carried out. In particular, we can view the dimensions of the vector space as attributes of the concept, and experimentally determined attribute importance as the weighting on that dimension. The compositional account and the distributional account are linked by the fact that they share the common structure of a \emph{compact closed category}. This allows the compositional rules of the grammar to be applied in the vector space model to map syntactically well-formed strings of words into one shared meaning space.

In the remainder of this section, we firstly describe pregroup grammar, used for the compositional rules. We then how the reductions of the grammar may be mapped to  finite dimensional vector spaces, and go on to give an example sentence reduction. We then describe a further type of structure, Frobenius algebras. These have been used to model information flow in sentences using relative pronouns such as `which'. We may therefore use these structures to model the sentence `pet which is a fish'.

\subsection{Pregroup Grammars}
In order to characterise the composition of concepts, the model uses Lambek's pregroup grammar \citep{lambek1999}. A pregroup   $(P, \leq, \cdot, 1, (-)^l, (-)^r)$ is a partially ordered monoid $(P, \leq, \cdot, 1)$ where each element $p\in P$ has a left adjoint $p^l$ and a right adjoint $p^r$, that is,  the following inequalities hold:

\[
p^l\cdot p \leq 1 \leq p\cdot p^l \quad \text{ and } \quad p\cdot p^r \leq 1 \leq p^r \cdot p
\]

The pregroup grammar then uses atomic types, such as $n$, $s$, their adjoints $n^l$, $s^r$..., and composite types which are forming by concatenating atomic types and their adjoints. We use the type $s$ to denote a declarative sentence and $n$ to denote a noun. A transitive verb can then be denoted $n^r s n^l$. If a string of words and their types reduces to the type $s$, the sentence is judged grammatical. The sentence `James shoots pigeons' is typed $n~(n^r s n^l)~ n$, and can be reduced to $s$ as follows: 
\[
n~(n^r s n^l)~ n \leq 1\cdot s n^l n \leq 1 \cdot s \cdot 1 \leq s
\]

However, this symbolic reduction can also be expressed  graphically as follows:

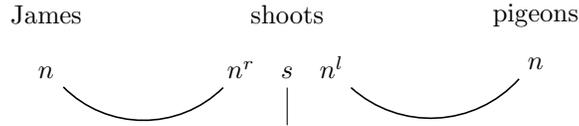
\begin{figure}[htbp]
\centering
\begin{tikzpicture}[node distance = 0.25cm and 2cm, text height = 1.5ex, bend angle = 45]
\node(james) {James};
\node(james_n) [below= of james] {$n$};
\node(shoots)[right= of james] {shoots}; 
\node(shoots_s)[below= of shoots] {$s$};
\node(end1)[below=0.5cm of shoots_s]{};
\node(n_shoots)[left=0.1cm of shoots_s] {$n^r$}
	edge [-, bend left, semithick] (james_n);
\node(shoots_n)[right=0.1cm of shoots_s] {${n^l}$};
\node(pig)[right=of shoots] {pigeons}; 
\node(pig_n)[below=0.1cm of pig] {$n$}
	edge [-, bend left, semithick] (shoots_n);

\draw[-] (shoots_s) -- (end1);
\end{tikzpicture}
%\caption{Reduction using the graphical calculus}
%\label{fig:reduct}
\end{figure}
In this graphical representation, the elimination of types by means of the inequalities $n \cdot n^r \leq 1$ and $n^l \cdot n \leq 1$ is represented by a `cup' while the fact that the type $s$ is retained is represented by a straight wire.

\subsection{Grammatical Reductions in Vector Spaces}
We give here a brief description of how the reductions of the pregroup grammar  may be mapped into the category of vector spaces. For full details, see \cite{coecke2010}.

The mapping uses the fact that both pregroups and vector spaces are instances of compact closed categories. We map the atomic types $n$, $s$ of the pregroup grammar to vector spaces $N$, $S$. In the category of vector spaces over $\mathbb{R}$, only one adjoint $N^*$ of $N$ exists and it is isomorphic to $N$, and hence $n^r$ and $n^l$ also both map to $V$. Similarly $s^r$, $s^l$ map to $S$.

The concatenation of types to form composites is mapped to the tensor product $\otimes$. So the sentence `James shoots pigeons', is typed $n n^r s n^l n$ in the pregroup grammar, and is represented in the tensor product of vector space $N \otimes N \otimes S \otimes N \otimes N$.

The reductions $\epsilon^r :n n^r \leq 1$, $\epsilon^l: n^l n \leq 1$ are mapped to the linear extension to the tensor product of the inner product:
\[
\epsilon: N\otimes N \rightarrow \mathbb{R}::\sum_{ij} c_{ij}\overrightarrow{v}_i \otimes \overrightarrow{w}_j \mapsto \sum_{ij} c_{ij}\braket{\overrightarrow{v}_i|\overrightarrow{w}_j} 
\]
and type introductions $\eta^r: 1 \leq n^r n$, $\eta^l: 1 \leq n n^l$ are implemented as
\[
\eta : \mathbb{R} \rightarrow N \otimes N :: 1 \mapsto \sum_i \overrightarrow{e}_i \otimes \overrightarrow{e}_i
\]

These are all presented mathematically rigorously in \cite{coecke2010} and an introduction to relevant category theory given in \cite{coecke2011}.

Now, individual words are assigned meanings within the vector space. This might be implemented via statistical analysis of text corpora, or attribute weightings elicited in an experiment. Nouns are represented as vectors in a single space $N$, whereas other word types are represented in larger spaces. For example, transitive verbs are typed $n^r s n^l$ and are therefore represented in the rank 3 space $N \otimes S \otimes N$. The reductions of the pregroup are implemented as inner products, and if the sentence is grammatical, one sentence vector $\overrightarrow{s} \in S$ should result. Meanings of sentences may then be compared by computing the cosine of the angle between their vector representations. So, if sentence $A$ has vector representation $\overrightarrow{s}_A$ and sentence $B$ has representation $\overrightarrow{s}_B$, their degree of synonymy is given by: 
\begin{equation}
\label{eq:sim}
\text{sim}(A, B) = \frac{\overrightarrow{s}_A \cdot \overrightarrow{s}_B}{||\overrightarrow{s}_A|| || \overrightarrow{s}_B||}
\end{equation}

We now give an example sentence reduction.

\subsubsection{An example sentence}
To express the sentence `James shoots pigeons', we firstly give types to the words in the sentence, so as before we have $n (n^r s n^l) n$. The reduction of the sentence is $\epsilon^r \otimes 1_s \otimes \epsilon^l$. Moving to the vector space, we  represent  nouns within a vector space $N$, sentences within a vector space $S$. A transitive verb is then represented in the tensor product space $N \otimes S \otimes N$.

So if we represent $\overrightarrow{shoots}$ by:  
\[
\overrightarrow{shoots} = \sum_{ijk} c_{ijk} \overrightarrow{e}_i \otimes \overrightarrow{s}_j \otimes \overrightarrow{e}_k
\]
then we have:
\begin{align*}
\overrightarrow{\text{James shoots pigeons}} &= \epsilon_N \otimes 1_S \otimes \epsilon_N(\overrightarrow{{James}} \otimes \overrightarrow{{shoots}} \otimes \overrightarrow{{pigeons}})\\
& = \sum_{ijk} c_{ijk} \braket{\overrightarrow{James}|\overrightarrow{e}_i} \otimes \overrightarrow{s}_j \otimes \braket{\overrightarrow{e}_k|\overrightarrow{pigeons}}\\
& = \sum_j\sum_{ik} c_{ijk} \braket{\overrightarrow{James}|\overrightarrow{e}_i} \braket{\overrightarrow{e}_k|\overrightarrow{pigeons}}\overrightarrow{s}_j
\end{align*}

Strings of words may also reduce to other types, such as nouns. An adjective can be given the type $n n^l$, and then a phrase such as `red car' is typed $n n^l n \leq n$. Another type of word that has been well studied is the possessive pronoun. In \cite{sadrzadeh2013}, the authors analyse the possessive pronoun as utilising a Frobenius algebra.

\subsection{Frobenius Algebras}
We state here how a Frobenius algebra is implemented within a vector space over $\mathbb{R}$. For a mathematically rigorous presentation see \cite{sadrzadeh2013}. A vector space $V$ over $\mathbb{R}$ with a fixed basis $\{\overrightarrow{v}_i\}_i$ has a Frobenius algebra given by:
\[
\Delta::\overrightarrow{v}_i \mapsto \overrightarrow{v}_i \otimes \overrightarrow{v}_i \quad \iota :: \overrightarrow{v}_i \mapsto 1 \quad \mu :: \overrightarrow{v}_i \otimes \overrightarrow{v}_i \mapsto \delta_{ij} \overrightarrow{v}_i \quad \zeta :: 1 \mapsto \sum_i \overrightarrow{v}_i
\]
 This algebra is commutative, so for the swap map $\sigma: X \otimes Y \rightarrow Y\otimes X$, we have $\sigma \circ \Delta = \Delta$ and $\mu \circ \sigma = \mu$. It is also special so that $\mu \circ \Delta = 1$. Essentially, the $\mu$ morphism amounts to taking the diagonal of a matrix, and $\Delta$ to embedding a vector within a diagonal matrix. This algebra may be used to model the flow of information in noun phrases with relative pronouns.

\subsubsection{An example noun phrase}
In \cite{sadrzadeh2013}, the authors describe how the subject and object relative pronouns may be analysed. We describe here the subject relative pronoun. The phrase `James who shoots pigeons' is a noun phrase; it describes James. The meaning of the phrase should therefore be James, modified somehow. The word `who' is typed $n^r n s^l n$, so the sentence `James who shoots pigeons' may be reduced as follows:

\begin{figure}[htbp]
\centering
\begin{tikzpicture}[node distance = 0.25cm and 2cm, text height = 1.5ex, bend angle = 90]
\node(james) {James};
\node(james_n) [below= of james] {$n$};
\node(who) [right= of james]{who};
\node(nr_who)[below left= 0.25cm and 0.05cm of who] {$n^r$}
	edge [-, bend left, semithick] (james_n);;
\node(end3)[below=1cm of nr_who]{};
\node(n_who)[right=0.1cm of nr_who] {$n$};
\node(end1)[below=0.75cm of n_who]{};
\node(end4)[below=1cm of n_who]{};
\node(who_s)[right=0.1cm of n_who] {$s$};
\node(who_n)[right=0.1cm of who_s] {$n$};
\node(shoots)[right= of who] {shoots}; 
\node(shoots_s)[below= of shoots] {$s$}
	edge [-, bend left, semithick] (who_s);
\node(n_shoots)[left=0.1cm of shoots_s] {$n^r$}
	edge [-, bend left, semithick] (who_n);
\node(shoots_n)[right=0.1cm of shoots_s] {${n^l}$};
\node(pig)[right=of shoots] {pigeons}; 
\node(pig_n)[below=0.1cm of pig] {$n$}
	edge [-, bend left, semithick] (shoots_n);
\draw[-] (n_who) -- (end1);
\end{tikzpicture}
%\caption{Reduction of `James who shoots pigeons'}
%\label{fig:jsp}
\end{figure}

\cite{sadrzadeh2013} go on to show that this may be reduced to 
\[
(\mu_N \otimes \iota_S \otimes \epsilon_N)(\overrightarrow{James} \otimes \overrightarrow{shoots} \otimes \overrightarrow{pigeons})
\]
This gives the result:
\begin{equation}
\label{eq:relpron}
\overrightarrow{\text{James who shoots pigeons}} = \overrightarrow{James}\odot (\underline{shoots} \times \overrightarrow{pigeons})
\end{equation}
where $\underline{shoots}$ is a matrix representing the verb `shoot', $\times$ refers to matrix multiplication and $\odot$ refers to elementwise multiplication.

We have given a brief overview of the key aspects of the model of meaning. These ideas will now be used to give an account of the `pet fish' phenomenon, firstly by recognising that `pet' in `pet fish' functions as an adjective, and secondly by analysing the expression `pet which is a fish'.

\section{A Compositional Distributional Account of the `Pet Fish' Phenomenon}
\label{sec:petfish}
We will use the compositional distributional model of meaning to give an account of the `pet fish' problem. The fuzzy-set theoretical view of the problem sees the types of the words `pet' and `fish' as the same, namely that both should be viewed as nouns and the word `pet fish' formed from their intersection. However, the word `pet' in this context is undeniably an adjective. Within pregroup grammar, it is typed as $n n^l$, and therefore within the model, this is viewed as a matrix $\underline{pet} = \sum_{ij} p_{ij} \overrightarrow{e}_i \otimes \overrightarrow{e}_j$. The meaning of `pet fish' is therefore 

\[
\overrightarrow{\text{pet fish}} = \sum_{ij} p_{ij} \overrightarrow{e}_i \braket{\overrightarrow{e}_j | \overrightarrow{fish}}
\]

The problem is essentially that an item such as a goldfish may not be a typical `fish', nor a typical `pet', but a very typical `pet fish'. To measure this within our model, we use the cosine similarity of meaning vectors $sim(\overrightarrow{A}, \overrightarrow{B})$ given in equation (\ref{eq:sim}) as a proxy for typicality. We may justify this by remarking that typicality may be characterised as as a function of similarity. Given two word vectors, e.g. $\overrightarrow{dog}$ and $\overrightarrow{pet}$, the vector representing `dog' should have a higher similarity to the concept `pet' than, for example `spider' should, i.e. we should have:
\begin{align*}
sim(\overrightarrow{dog}, \overrightarrow{pet}) &= \frac{\overrightarrow{dog}\cdot\overrightarrow{pet} }{||\overrightarrow{dog}||||\overrightarrow{pet}||}\\
& > \frac{\overrightarrow{spider}\cdot\overrightarrow{pet} }{||\overrightarrow{spider}||||\overrightarrow{pet}||} \\
& = sim(\overrightarrow{spider}, \overrightarrow{pet})
\end{align*}

We now examine the effect of concept combination on cosine similarity, and see that the `pet fish' phenomenon is reproduced.

\subsection{Creating Adjectives}
\label{sec:adj}
Adjectives, verbs, adverbs and various other grammatical types may be viewed as operators. An adjective, in particular, is a matrix in $N \otimes N$, and the application of an adjective to a noun within the framework is matrix multiplication of the noun by the adjective. It is simple to craft a matrix $\underline{pet}$ that transforms the vector $\overrightarrow{fish}$ such that:
\begin{align*}
sim(\underline{pet}\times\overrightarrow{fish}, \overrightarrow{goldfish}) > sim(\overrightarrow{fish}, \overrightarrow{goldfish})
\end{align*}
where $\times$ refers to matrix multiplication.

However, this process may be simplified even further. \cite{kartsaklis2013} characterise an $n$-ary operator by the sum of the words it takes as arguments. For example `pet' is $\sum_i \overrightarrow{n_i}$ where the $\overrightarrow{n_i}$ are words that co-occur with `pet' such as `dog', `cat', `spider' and so forth. Each word may occur more than once in this sum to give a frequency calculation. This gives a tensor whose rank is one less than the desired rank. Therefore, \cite{kartsaklis2013} suggest expanding the rank using the Frobenius copy operator $\Delta$. The meaning of an adjective-noun combination such as `pet fish' may then be calculated as

\[
\overrightarrow{\text{pet fish}} = \sum_{ij} p_{ij} \overrightarrow{e}_i \braket{\overrightarrow{e}_j | \overrightarrow{fish}}
\]

If the Frobenius copy operation and the inner product are both carried out with respect to the same basis, we obtain 
\[
\overrightarrow{\text{pet fish}} = \overrightarrow{pet_{adj}}\odot \overrightarrow{fish_{noun}}
\]

Although the vectors $\overrightarrow{pet_{adj}}$ and $\overrightarrow{pet_{noun}}$ are likely to be similar, they will not be identical. This allows the combinations `pet fish' and `fishy pet' to be different. We will see that this method of forming adjectives can reproduce the qualitative phenomena required, and it is notable that we can do this without expanding the rank of the data space.

Another way of expressing the combination of the two concepts `pet' and  `fish' is by the extended `fish which is a pet', or `pet which is a fish'. In the experiments from \cite{hamp1987, hamp1988a, hamp1988b}, this is how the two combinations are phrased. One of the key findings from these experiments was typicality of an item to each ordering is not identical. In order to express `fish which is a pet', we use the verb `to be' as a transitive verb. These are typed as $N \otimes S \otimes N$ and can also be expressed as a sum of their arguments, i.e. as $\sum_{ij} \overrightarrow{n}_i \otimes \overrightarrow{n}_j$ where $\overrightarrow{n}_i$ is the subject of the verb and $\overrightarrow{n}_j$ the object. Here we have discarded the sentence type $S$. In the  treatment of the relative pronoun `which', the sentence type is also discarded, and therefore the verb matrix can be used directly. We can express the combination `fish which is a pet' analogously to the expression given in equation \ref{eq:relpron}:
\[
\overrightarrow{\text{fish which is a pet}} = \overrightarrow{fish}\odot (\underline{is} \times \overrightarrow{pet})
\]
where $\underline{is}$ is the matrix of the verb `to be'.

In the next section we show that these two methods of combination give us some of the qualitative  properties we need for the `pet fish' phenomenon.

\subsection{Toy Model Using Adjective Vectors}
This toy model applies the way of forming an adjective as a sum of noun vectors. Suppose that the nouns `dog', `cat', `goldfish', `shark', `pet' and `fish' have attribute weights as presented in table \ref{tab:toy_sup}.  Each attribute is weighted as given by the entries in the table. These weights are hypothetical weights, but could be elicited from humans in an experiment, or from text corpora. We can form the adjective `pet' by forming a superposition of `dog', `cat', and `goldfish':

\[
\overrightarrow{pet_{adj}} = \overrightarrow{dog} + \overrightarrow{cat} + \overrightarrow{goldfish} = [2.5, 0.6, 1, 1, 2.7]^\top
\] 

The cosine similarity of each animal to the nouns `pet', `fish', and `pet fish' are shown in table \ref{tab:cos_sup}. The qualitative phenomenon that a goldfish is a better example of a pet fish than of either a pet or a fish is exhibited.
\begin{table}[htbp]
\centering
\caption{List of concepts and attributes for adjective vector model}
\label{tab:toy_sup}
\begin{tabular}{| l | l | l | l | l | l | l | }
  \hline 
				&pet		&fish	&goldfish	&cat	 & dog	& shark   \\   
\hline          
  cared-for 		&1 		& 0.2	& 0.7		&0.9	 &0.9	& 0 \\
  vicious 		&0.2		& 0.8	& 0			&0.2	 &0.4 & 1 \\
  fluffy 		&0.7		& 0		& 0			&0.9	 &0.7 & 0  \\
  scaly 			&0.2		& 1		& 1			&0	&0	& 1\\
  lives in the sea	&0		& 0.8		& 0			&0	&0	& 1\\
  lives in house	&0.9		& 0.3	& 0.9		&0.9	 & 0.9	& 0\\
  \hline  
\end{tabular}
\end{table}

\begin{table}[htbp]
\centering
\caption{Cosine similarity for adjective vector model}
\label{tab:cos_sup}
\begin{tabular}{| l | l | l | l | l | l | l |}
  \hline 
							&goldfish	&cat	 	& dog	& shark  \\   
\hline          
  pet (noun) 				& 0.7309		&0.9816	&0.9809 	& 0.1497 \\
  fish 						& 0.5989		&0.2500 &0.3292	& 0.9670 \\
  pet (adj) fish 			& 0.9377		&0.5524	&0.6197	& 0.5861 \\
  \hline  
\end{tabular}
\end{table}
\subsection{Toy Model Using Relative Pronouns}
We give here a toy model for the composition of two concepts using Frobenius multiplication, and show that the qualitative attributes we require are exhibited. Suppose we again have attributes as listed in the rows of table \ref{tab:toy_sup} for the concepts as given in the columns.

Weights for the matrix for the verb `to be' are given in table \ref{tab:toy_is} below. These weights express the extent to which the attributes co-occur. For example, the extent to which something is  vicious is also cared-for is given weight 0.02. We assume that the extent to which an attribute is itself, for example, the extent to which a vicious thing is vicious, is greater than the extent to which it is anything else. Hence the diagonal is emphasised. Ideally, the verb `to be' would interact directly with the attribute space as does the relative pronoun, as this will form a future line of enquiry.
\begin{table}[htbp]
\centering
\caption{Matrix of weights for the verb `to be'}
\label{tab:toy_is}
\begin{tabular}{| l | l | l | l | l | l | l |}
  \hline 
				&cared-for	&vicious 	&fluffy	&scaly	& water & house   \\   
\hline          
  cared-for 		&1			& 0.02		& 0.08	& 0.03	& 0.02 & 0.09 \\
  vicious 		&0.02 	& 1 				&0.05	&0.06 	& 0.05 &0.02\\
  fluffy 		&0.08 	&0.05			&1		&0		&0		&0.09  \\
  scaly 			&0.03 & 0.06 &  0& 1&     0.05& 0.02\\
  lives in the sea	&0.02 & 0.05 & 0& 0.05&   1&   0\\
  lives in house	&0.09 & 0.02 & 0.09 & 0.02 & 0& 1\\
  \hline  
\end{tabular}
\end{table}

Using this representation of the verb `to be' and the grammatical structure of relative pronouns, we have the following: 
\[
\overrightarrow{\text{pet which is a fish}} =  \overrightarrow{pet}\odot (\underline{is}\times \overrightarrow{fish}) = [0.29,   0.18,  0.06, 0.22, 0.22, 0.35]^\top
\]
and 
\[
\overrightarrow{\text{fish which is a pet}} =  \overrightarrow{fish}\odot (\underline{is}\times \overrightarrow{pet}) = [0.23, 0.24, 0,    0.27,    0.27,    0.32]^\top
\]
 The similarity of each of the vectors for `goldfish', `cat', and `shark' to the concepts `pet', `fish', `pet which is a fish' and `fish which is a pet'  is given in table \ref{tab:frob_sim}. We see that the similarity of `goldfish' to `pet which is a fish' and `fish which is a pet' is higher than its similarity to either `pet' or `fish'. In addition, similarity of each animal to `pet which is a fish' is not identical to its similarity to `fish which is a pet'.
\begin{table}[htbp]
\centering
\caption{Cosine similarity for relative pronoun model}
\label{tab:frob_sim}
\begin{tabular}{| l | l | l | l | l | l | }
  \hline 
							&goldfish	&cat		& shark  \\   
\hline          
  pet 						& 0.7309		&0.9816	& 0.1497 \\
  fish 						& 0.5989		&0.2500	& 0.9670 \\
  pet which is a fish 		& 0.8999		&0.7783	& 0.4467 \\
  fish which is a pet 		& 0.8898		&0.6540	& 0.5730 \\
  \hline  
\end{tabular}
\end{table}

\section{Modelling Attribute Combination}
\label{sec:exp}
In \cite{hamp1987}, Hampton collects data on the importance of attributes in concepts and their combination. Concepts are considered in pairs that are related to some degree, for example `Pets', and `Birds'. Six pairs are considered in total, detailed below. Participants are asked to generate attributes for each concept and for their conjunctions, where conjunction in this case is rendered as `Pets which are also Birds', or `Birds which are also Pets'. For example, attributes such as: `lives in the house', `is an animal', `has two legs', are generated for `Pets', `Birds'. For each pair of concepts and their conjunction, attributes that had been generated by at least 3 out of the 10 participants were collated. Participants were then asked to rate the importance of each attribute to each concept and to each conjunction. Importance ratings were made on a 7 point verbal scale ranging from `Necessarily true of all examples of the concept' to `Necessarily false of all examples of the concept'. Numerical ratings were subsequently imposed ranging from 4 to -2 respectively.

The question then arises of how the importance of attributes in the conjunction of the concepts is related to the importance of attributes in the constituent concepts. Key phenomena are that conjunction is not commutative, that inheritance failure can occur (i.e., an attribute that is important in one of the concepts is not transferred to the conjunction), that attribute emergence can occur, where an attribute that is important in neither of the conjuncts becomes important in the conjunction, and further, that necessity and impossibility are preserved. In order to model this data, Hampton uses a multilinear regression.

We use the importance values for each attribute and their conjunction to determine a set of weights for the verb `to be', which, when substituted in to the phrase `A which is a B', or `B which is an A', provides the appropriate attribute weights. We require a matrix $\underline{is}$ such that:
\[
\overrightarrow{A} \odot (\underline{is} \times \overrightarrow{B}) = \overrightarrow{AB}
\]
and 
\[
\overrightarrow{B} \odot (\underline{is} \times \overrightarrow{A}) = \overrightarrow{BA}
\]
where $\overrightarrow{AB}$ stands for the combination `A which is a B' and vice versa. The importance values are rated on a scale $4, 3, 2, 1, -1, -2$. We map these into a $[0,1]$ interval by $r \mapsto (r + 2)/6$, where r is the relevant rating.

To fit parameters, we use MATLAB's \texttt{fmincon} with the active-set algorithm and constraint that verb entries must be greater than 0. Results are reported in tables \ref{tab:cos1} and \ref{tab:cos2}. F-HA stands for `Furniture and Household Appliances', F-P `Foods and Plants', W-T `Weapons and Tools', B-D `Buildings and Dwellings', M-V `Machines and Vehicles', B-P `Birds and Pets'.
\begin{table}[htbp]
\centering
\caption{Cosine similarity measure for multilinear regression (MLR) and for the compositional distributional (CD) model, for the ordering `$A$ which is a $B$'}
\label{tab:cos1}
\begin{tabular}{| l | l | l | l | l | l | l |}
  \hline 
	Cosine similarity			&F-HA	&F-P		& W-T & B-D & M-V & B-P  \\   
\hline          
  MLR			& 0.9905   & 0.9944  &  0.9791  &  0.9839  &  0.9948  &  0.9727\\
  CD 			& 0.9996  &  1.0000  &  0.9997  &  0.9996  &  1.0000  &  0.9999 \\
  \hline  
\end{tabular}
\end{table}
\begin{table}[htbp]
\centering
\caption{Cosine similarity measure for multilinear regression (MLR) and for the compositional distributional (CD) model, for the ordering `$B$ which is a $A$'}
\label{tab:cos2}
\begin{tabular}{| l | l | l | l | l | l | l |}
  \hline 
	Cosine similarity			&F-HA	&F-P		& W-T & B-D & M-V & B-P  \\   
\hline          
  MLR			& 0.9898  &  0.9965  &  0.9930  &  0.9945  &  0.9948  &  0.9673\\
  CD		& 0.9997  &  1.0000  &  0.9996  &  0.9997  &  1.0000  &  0.9999 \\
  \hline  
\end{tabular}
\end{table}
Modelling the combination of the pairs of concepts using the grammatical attributes of the phrase `A which is a B' allows for a greater accuracy in modelling the conjunction. In the tables above, we have used the same matrix for the verb `is' in each ordering. It is unsurprising that we are able to obtain a good fit to the data, since we are using $k^2$ weights in the verb `to be' for $2k$ datapoints, and the results we show here are therefore not of any statistical significance. However, by using the verb `to be' in this way, we are able to take account of how attributes interact with one another. Further work will examine how the matrices thus obtained reflect Hampton's findings regarding the necessity and impossibility of attributes, attribute emergence and inheritance failure.

\section{Discussion}
\label{sec:disc}
We have described a compositional distributional model of meaning \citep{coecke2010}, which utilises grammar to describe how the meaning of a compound arises from the meaning of its parts. This account of meaning is inherently compositional, and importantly, the meanings of composites inhabit the same space as their constituents, so that the inner product may be used to compare concepts directly. We apply this model to the `pet fish' problem, describing how the phrases `pet fish' and `pet which is a fish' may be modelled within the formalism, and giving two illustrative models which show how the qualitative phenomena are naturally produced. We go on to model a set of data from \cite{hamp1987}. This highlights how attributes in concepts interact, via the verb `to be'.

The approach we have outlined contrasts with approaches in the quantum cognition literature departing from an assumption of non-compositionality. The claim of non-compositionality in the literature refers to the fact that human concept combination and judgements cannot be modelled using classical probability. Instead, we take a semantic approach.  Within the model of meaning that we describe, the meaning of a sentence is rendered as a vector. The meaning of individual words in the sentence or noun phrase are given by vectors or tensors, and a method for combining them is specified. We therefore view the meaning of a complex expression as being exactly specified by the meaning of its parts and of how they are combined. The `non-compositional' phenomena described may well have a description that is compositional when both meaning and grammar are taken into account, and in particular we have argued that the `pet fish' example may be viewed as compositional.

Further work will extend the account to other cognitive phenomena. We can straightforwardly apply the modelling of `pet fish' to account for the conjunction fallacy \cite{tversky1983}, since this is an example of overextension. Another oft-cited phenomenon is the asymmetry of similarity judgements \citep{tversky1977}, in which, for example, Korea is judged more similar to China than China is to Korea. There are two approaches we might take to modelling this. Firstly, as detailed in \cite{coecke2010}, we can choose a graded truth-theoretic space as the sentence space. Then, the sentence `Korea is similar to China' can be modelled as mapping to a higher value than `China is similar to Korea'. 

Alternatively, a fuller account of the ways in which concepts interact can be developed. Whilst synonymy is useful as a proxy for membership, and of course in comparing meaning, we must develop measures that allow the description of various relationships between concepts. \cite{balkir2014} uses the non-symmetric measure of relative entropy to characterise hyponymy, which is related to the idea of membership in a concept. Hyponymy is a stricter notion than membership, however, and therefore we need to generalise this model. Other relationships are typicality and meronymy, where a concept forms part of another concept, such as `finger' to `hand'.

We will further investigate evidence for these type of phenomena in text corpora, and different ways of modelling adjectives. Given a large enough number of dimensions, it is possible to create a matrix for an adjective that can exactly recreate the meaning vector for an adjective-noun combination. It would be interesting to look at the existence of this type of phenomena when the context, as expressed by a choice of basis vectors, is specified, rather than being, for example, the most commonly used words in the corpus.

Another area for research is the interaction of ambiguous concepts. These ideas are investigated in \cite{bruza2013}, in which the authors elicit similarity judgements on novel combinations of ambiguous words such as `apple chip' - here, `apple' could be be interpreted as either a fruit or a computer brand, and `chip' as either food or hardware. The authors define non-compositionality as the failure of the interpretation of the combination of two concepts such as `apple chip' to be modelled as a joint probability distribution over the interpretations of the two constituent concepts. Within the compositional distributional model of meaning that we have described, ambiguity in word meanings can be modelled by the use of density matrices, and this ambiguity interacts with other words in the sentence which may serve to disambiguate the word \citep{piedeleu2015}. It would be interesting to model the phenomena found in \cite{bruza2013} within this framework. 

Finally, the role of the verb `to be' should be investigated. This verb should have a functional role, as do relative pronouns, as well as a distributionally determined meaning. 

\section*{Acknowledgements}
Martha Lewis gratefully acknowledges support from EPSRC (grant EP/I03808X/1) and AFOSR grant Algorithmic and Logical Aspects when Composing Meanings. Many thanks to James Hampton for use of datasets.

\bibliographystyle{plainnat}
%\bibliography{../../phd} 

\end{document}